%% file: Bioinformatics.tex
\documentclass[unnumsec,webpdf,contemporary,large]{oup-authoring-template}%
\usepackage{times}
\usepackage{soul}
\usepackage{url}
\usepackage[hidelinks]{hyperref}
\usepackage[utf8]{inputenc}
\usepackage{graphicx}
\usepackage{amsmath}
\usepackage{amsthm}
\usepackage{booktabs}
\usepackage{algorithm}
\usepackage{amsfonts,amssymb}

\usepackage{algpseudocode} 
\usepackage[normalem]{ulem}

\graphicspath{{Fig/}}

\theoremstyle{thmstyleone}%
\theoremstyle{thmstyletwo}%
\theoremstyle{thmstylethree}%

\begin{document}

\journaltitle{Bioinformatics}
\DOI{DOI HERE}
\copyrightyear{2023}
\pubyear{2019}
\access{Advance Access Publication Date: Day Month Year}
\appnotes{Paper}

\firstpage{1}

%\subtitle{Subject Section}

\title[Short Article Title]{Structure-aware Protein Self-supervised Learning}

\author[1, 2]{Can (Sam) Chen}
\author[3]{Jingbo Zhou}
\author[4]{Fan Wang}
\author[1]{Xue Liu}
\author[5]{Dejing Dou}

\authormark{Bioinformatics, 2023}

\address[1]{\orgname{McGill University}, \orgaddress{ \country{Canada}}}
\address[2]{\orgname{MILA - Quebec AI Institute}, \orgaddress{ \country{Canada}}}
\address[3]{\orgname{Baidu Research}, \orgaddress{ \country{China}}}
\address[4]{\orgname{Baidu Inc.}, \orgaddress{ \country{China}}}
\address[5]{\orgname{BCG X}, \orgaddress{ \country{China}}}

%\corresp[$\ast$]{Corresponding author. \href{email:email-id.com}{email-id.com}}

\received{Date}{0}{Year}
\revised{Date}{0}{Year}
\accepted{Date}{0}{Year}

%\editor{Associate Editor: Name}

%\abstract{
%\textbf{Motivation:} .\\
%\textbf{Results:} .\\
%\textbf{Availability:} .\\
%\textbf{Contact:} \href{name@email.com}{name@email.com}\\
%\textbf{Supplementary information:} Supplementary data are available at \textit{Journal Name}
%online.}

\abstract{
\textbf{Motivation.} Protein representation learning methods have shown great potential {to} many downstream tasks {in biological applications}. 
{A few recent studies have demonstrated that the self-supervised learning is a promising solution to addressing insufficient labels of proteins, which is a major obstacle to {effective} protein representation learning.}
However, existing protein representation learning is usually pretrained on protein sequences without considering the important protein structural information.
\\
\textbf{Results.} In this work, we propose a novel structure-aware protein self-supervised learning method to effectively capture structural information of proteins.
In particular, a graph neural network (GNN) model is pretrained to preserve the protein structural information with self-supervised tasks from a pairwise residue distance perspective and a dihedral angle perspective, respectively.
Furthermore, we propose to leverage the available protein language model pretrained on protein sequences to enhance the self-supervised learning.
Specifically, we identify the relation between the sequential information in the protein language model and the structural information in the specially designed GNN model via a novel pseudo bi-level optimization scheme. 
We conduct experiments on three downstream tasks: the binary classification into membrane/non-membrane proteins, the location classification into $10$ cellular compartments, and the enzyme-catalyzed reaction classification into $384$ EC numbers, and these experiments verify the effectiveness of our proposed method.
\\
\textbf{Availability and Implementation.}
The Alphafold2 database is available in \url{https://alphafold.ebi.ac.uk/}.
The PDB files are available in \url{https://www.rcsb.org/}.
The downstream tasks are available in \url{https://github.com/phermosilla/IEConv\_proteins/tree/master/Datasets}.
The code of the proposed method is available in \url{https://github.com/GGchen1997/STEPS_Bioinformatics}.
\\
\textbf{Contact.} Reach Jingbo Zhou from Baidu Research by zhoujingbo@baidu.com or Can (Sam) Chen from McGill and Mila - Quebec AI Institute by can.chen@mila.quebec.
}

% \boxedtext{
% \begin{itemize}
% \item Key boxed text here.
% \item Key boxed text here.
% \item Key boxed text here.
% \end{itemize}}

\maketitle

\input{Sec1_introduction}

\input{Sec2_method}
\input{Sec3_experiment}
\input{Sec4_relatedwork}
\input{Sec5_conclusion}

\input{Sec6_appendix}
\section{Acknowledgments}
We would like to thank the editorial team and the publisher of Bioinformatics for granting us a waiver of the open access fee for this publication. We appreciate their support in promoting open access and helping our research reach a wider audience.

\bibliographystyle{alpha}
\bibliography{reference}

%USE THE BELOW OPTIONS IN CASE YOU NEED AUTHOR YEAR FORMAT.
%\bibliographystyle{abbrvnat}
%\bibliography{reference}

\end{document}

%% file: Sec1_introduction.tex
%\textcolor{blue}{[MA: maybe change the title to Noise-free Top-K recommender via Bi-level Optimization.]}

\section{Introduction}

A variety of machine learning-based biological tasks heavily rely on effective protein representation learning, which aims to extract rich sequential and structural information of a protein into a high-dimensional vector. 
The learned protein representation can be used in many downstream tasks such as protein function annotation \cite{gligorijevic2021structure}, enzyme-catalyzed reaction prediction \cite{hermosilla2020intrinsic}, and protein classification \cite{almagro2017deeploc}. 
%
% \old{Recently protein representation learning, which exhibits %\old{great advantages}
% promising performance over feature engineering methods, has attracted lots of research attention.}
%
%\old{These methods models protein from different levels including primary structure, secondary structure and tertiary structure.}
%
{With this topic attracting lots of research attention recently, }
different neural network architectures are adopted to learn different levels of protein information {based on the labeled protein data via supervised learning.}
%\old{where LSTM-based and Transformer-based model}
For example, LSTMs~\cite{sonderby2015convolutional} are used to model the sequential information (i.e. primary structure of protein), and variants of graph neural networks~\cite{gligorijevic2021structure} and convolutional neural networks~\cite{hermosilla2020intrinsic} are used to model the structural information.
Though these deep learning-based models prove to be effective, one major obstacle to such approach is the lack of labeled data, which is much more severe than the one in computer vision and natural language processing areas since the wet-lab experiment on protein is quite  expensive.
%\todo{classification or function prediction} 
% \old{is the lack of annotated protein data, which is much more severe than the one in computer vision and natural language processing areas since the wet-lab experiment of a protein annotation is quite  expensive.}

%(i.e. secondary structure and/or tertiary structure) in protein.}  
%\old{LSTMs and Transformers are developed to model the sequential information in protein and graph neural network-based models are used to model the structure information in protein.}
%
%\old{These methods better capture protein feature than traditional methods in many aspects.}
%
%LSTM and transformer based methods mainly treat protein as a sequence of amino acids, which models the sequential 
%
%\old{Yet,} 
% \old{Though these deep learning-based models prove to be effective, one major obstacle of protein classification
% %\todo{classification or function prediction} 
% is the lack of annotated protein data, which is much more severe than the one in computer vision and natural language processing areas since the wet-lab experiment of a protein annotation is quite  expensive.} %\old{cost much than that of an image or a sentence.}
%

Inspired by the remarkable progress of self-supervised learning {in other domains}, 
%\old{in the field of natural language processing,} %\old{recent work performs} 
there are a few recent work to perform 
%\old{pretraining on protein datasets to learn a protein representation}
self-supervised learning for protein from a sequence perspective~\cite{bepler2019learning,rives2021biological,rao2019evaluating,elnaggar2020prottrans,rao2020transformer,vig2020bertology}.
%\todo{check if ssl}
%or a structure perspective.
%
%\todo{add line number, distance information and dihedral angle information}
%
These sequence-based pretraining methods treat every protein as a sequence of amino acids and use autoregressive or autoencoder methods to obtain the protein representation.
Although previous studies~\cite{rives2021biological,elnaggar2020prottrans} found such sequential pretrained protein language models can understand protein structures to some extent, 
these studies have not explicitly considered
%\zhou{these models cannot} 
modeling
%\zhou{3D} 
structural information of proteins. 
%\old{structure.}
%
%As shown in Figure~\ref{fig:protein_struc}, the protein backbone consists of consecutive units of $C_{\alpha}$–CO–NH–$C_{\alpha}$ with two dihedral angles: Phi ($\phi$) and Psi ($\psi$).
%\todo{structure determines function}
%
%Protein structure 
% \old{Protein structural information determines a wide range of protein properties~\cite{gligorijevic2021structure}, but is not explicitly utilized on self-supervised learning for proteins. }

%The distance information captures the global structure information and the dihedral angle information captures the local structure information \zhoucom{is there any citations for this one?}, both of which captures essential information of protein  \old{are neglected in pretraining. }

% \old{Although these protein representation learning methods achieve comparable performance in various tasks compared with feature engineering methods,} 

{Though protein structural information determines a wide range of protein properties~\cite{gligorijevic2021structure}, how to incorporate protein structural information into protein self-supervised learning is  overlooked. }
With the development of structural biology including cryo-EM~\cite{callaway2020revolutionary} and Alphafold2~\cite{jumper2021highly}, the availability of reliable protein structures
is increasing in recent years.
%
%\zhou{Yet, the availability of protein structures is increasing in recent years. Even the cryo-EM is still expensive, the breakthrough of Alphafold2 \cite{jumper2021highly} reveals reliable structure for many proteins based on sequences. } 
Thus, it is desirable to devise a new mechanism to explicitly incorporate protein structural information into self-supervised learning to boost the performance of {protein representation learning.} 
%\old{protein classification.} %\zhou{Yes, unlike to pretrain the model on massive structural molecules\cite{}, the proteins with knowing structure information is still very limited compared with the ones with knowing amino acid sequence. Thus,  pre-training a protein language model solely based on the  limited structural protein data may not be able to show superior performance compared with exiting protein language models.}
Meanwhile, the number of protein sequences {is still orders of magnitude larger} 
%\old{is still much larger} 
than the number of proteins with reliable protein structures.
%
%\todo{it is desirable to leverage the availabel protein lm}
%\todo{no 3D or 1D but structural and sequential}
Therefore, learning protein representation solely based on the limited number of structural protein data may not be able to show superior performance compared with existing protein language models. %pretrained on the large scale of protein sequences. %\old{prtrained protein language models.}
To this end, we propose a novel \textit{\textbf{ST}rucure-awar\textbf{E} \textbf{P}rotein
\textbf{S}elf-supervised Learning}~(\textbf{STEPS}) method.
This method can not only explicitly incorporate protein structural information into protein {representations}, but also leverage the existing protein language model to enhance protein representation learning.
%and connect the model training to the available protein language model via bi-level optimiz
%into protein modeling.
%
%To effectively 
% \old{Specifically, 
% %the local-level self-supervised task aims to predict the angle and the global-level self-supervised task aims to predict the distance between 
% %The self-supervised angle prediction task preserves local 3D information and the distance prediction task preserves global 3D information of protein.
% %
% we design a graph neural network~(GNN) to effectively model the protein structure.
% %
% The GNN model takes the masked protein structure as input and aims to reconstruct
% the pairwise residue distance and the dihedral angle.}
%
More specifically, we leverage a graph neural network (GNN) to model protein structure
%the modeling of protein 3D structure. 
and
%\old{two novel self-supervised learning tasks are proposed to
%incorporate the the distance information and the angle information into protein modeling.} 
propose two novel self-supervised learning tasks to incorporate the distance information and the angle information into protein {representation learning.} %\old{modeling. }
In particular, the GNN model takes the masked protein structure as input and aims to reconstruct
the pairwise residue distance information and the dihedral angle information respectively.

Furthermore,  we propose to leverage the available sequential protein language model pretrained on protein sequences~(named as protein LM for short) to empower the GNN model via a pseudo bi-level optimization scheme. 
%
%A straightforward solution for this purpose is to use the protein LM to obtain embedding vectors for all amino acids as input of structure-aware GNN. However, the disadvantage of this method is that the protein LM still cannot learn the structure information of the protein, resulting in degenerative performance in downstreaming tasks. In our SPM framework, we take the protein LM as  auxiliary input to train the structure-aware GNN which is a upper level optimization task. %by pre-defined self-supervised learning tasks (which is a upper level task optimization). 
%Meanwhile, the protein LM is also refined by the protein structure information which is a lower level optimization task. \zhoucom{for bi-level optimiation, there should two optimization objective function for different variables?}  Thus, to tackle the interplay between the sequential information and the structural information, we propose a bi-level optimization strategy to optimize the protein LM and the structure-aware GNN via self-supervised learning. In this way, the structural information is not only updated through the structure-aware GNN model but also via the protein LM model, which could better exploit the structural knowledge in the LM.
%
{This optimization scheme aims to effectively transfer the knowledge of the protein LM to the GNN model. 
%\sout{The insight is that we identify a relation between the protein LM and the GNN model and the relation is defined by the constraint between protein sequence and protein structure.}
%
The insight is that we identify the relation between the sequential information and the structural information by maximizing the mutual information between the sequential representation and the structural representation.
%
%Then a bi-level optimization scheme is devised to  exploit the sequential information in the protein LM by tackling the interplay between sequential information in the pretrained protein language model and structural information in the GNN model. 
Then a bi-level optimization scheme is devised to exploit the sequential information in the protein LM by leveraging its relation with the structural information in the GNN model. 
%The structure-aware GNN is updated under the relation, which can better exploit the sequential information in the protein LM.
%
We name this optimization process as
\textit{pseudo bi-level} optimization because we update the GNN model in the outer level, but finally keep the parameters of the protein LM fixed in the inner level to avoid distorting the protein LM.}
Experiments on several downstream tasks
%including the membrane/non-membrane protein classification, the location classification and the enzyme classification 
verify the effectiveness of STEPS. 

In summary, we make the following contributions:
\begin{itemize}
    \item To the best of our knowledge, we are the first to explicitly incorporate \textcolor{black}{finer} protein structural information into self-supervised learning. Two novel self-supervised tasks are proposed to capture the pairwise residue distance information and the dihedral angle information respectively.
    \item 
    %We adopt a pseudo bi-level optimization scheme where we identify the relation between the protein LM and the structure-aware GNN and leverage the relation to better exploit the sequential information in the protein LM
    We adopt a pseudo bi-level optimization scheme to exploit the sequential information in the protein LM.
    %after identifying the relation between the protein LM and the structure-aware GNN.

    %to  the relation between the LM and GNN.
    %
    %In this way, the knowledge in the pretrained LM can be better exploited to improve the protein pretraining.
    %keep the consistency between sequential information and structural information by leveraging the available protein language model.
    %introduce bi-level optimization to fuse structural level and sequential level protein information.
    \item We conduct various supervised downstream tasks to verify the effectiveness of {STEPS}.
\end{itemize}
%
%The rest of paper is organized as follows.
%
%In 

% 1. fix the languge model prior and integrate the gnn learning under the LM constraint;
% small datasets of limited protein are not enough for fine-tuning.
% Since the embeddings already reveal sth, we do not intend to change it.

% 2. hinge loss to the two residues.

% 3. Why do not finetune? what is the difference bewteen this setting and previous setting?
% Previous pretrained model have everything needed in their their self-supervised tasks, like in Bert.
% Protein LM do not use strucure info in their self-supervised tasks.

%% file: Sec2_method.tex
\section{Preliminaries}
%In this section, we first introduce some important biological concepts related to proteins.
In this section, we first introduce preliminary concepts and some basic notations used in this paper.
%\old{some important biological concepts related to proteins.}

\noindent \textbf{Protein sequence.}
Each protein $\rm{S}(V, \mathcal{E})$ is a sequence of residues linked by peptide bonds.
Here $\mathcal{V}$ represents the set of $L$ residues in the sequence and $\mathcal{E} \subseteq {V} \times {V} $ describes the $L-1$ peptide bonds.
{The protein sequence is mainly composed of twenty different types of amino acids where each unit
%of the protein sequence 
is commonly named as residue after being joined by peptide bonds. }

%
%\old{There are commonly twenty \old{kinds of} amino acids in protein sequences and 
%
%Like language consists of words, protein sequence is composed of these amino acids.
%
%The amino acids are linked by peptide bonds and then form a linear sequence of the protein backbone.
%
%We use the full name(e.g., Proline), or the 3-letter abbreviation (Pro), or the single-letter code (P) to represent an amino acid.
%
%each unit 
%\todo{explain what is side chain; replace side chain with unit}
%is one \old{among the twenty kinds of} \zhou{of these twenty} amino acids.}
%
%\old{The sequence length $L$ is uncertain, which results in a huge combinatorial space of protein sequences. } 
%The sequence length $L$ is uncertain, which results in a huge combinatorial space of protein sequences.
%\todo{why you delete this?}

%
\begin{figure}
    \centering
    \includegraphics[width=1.0\columnwidth]{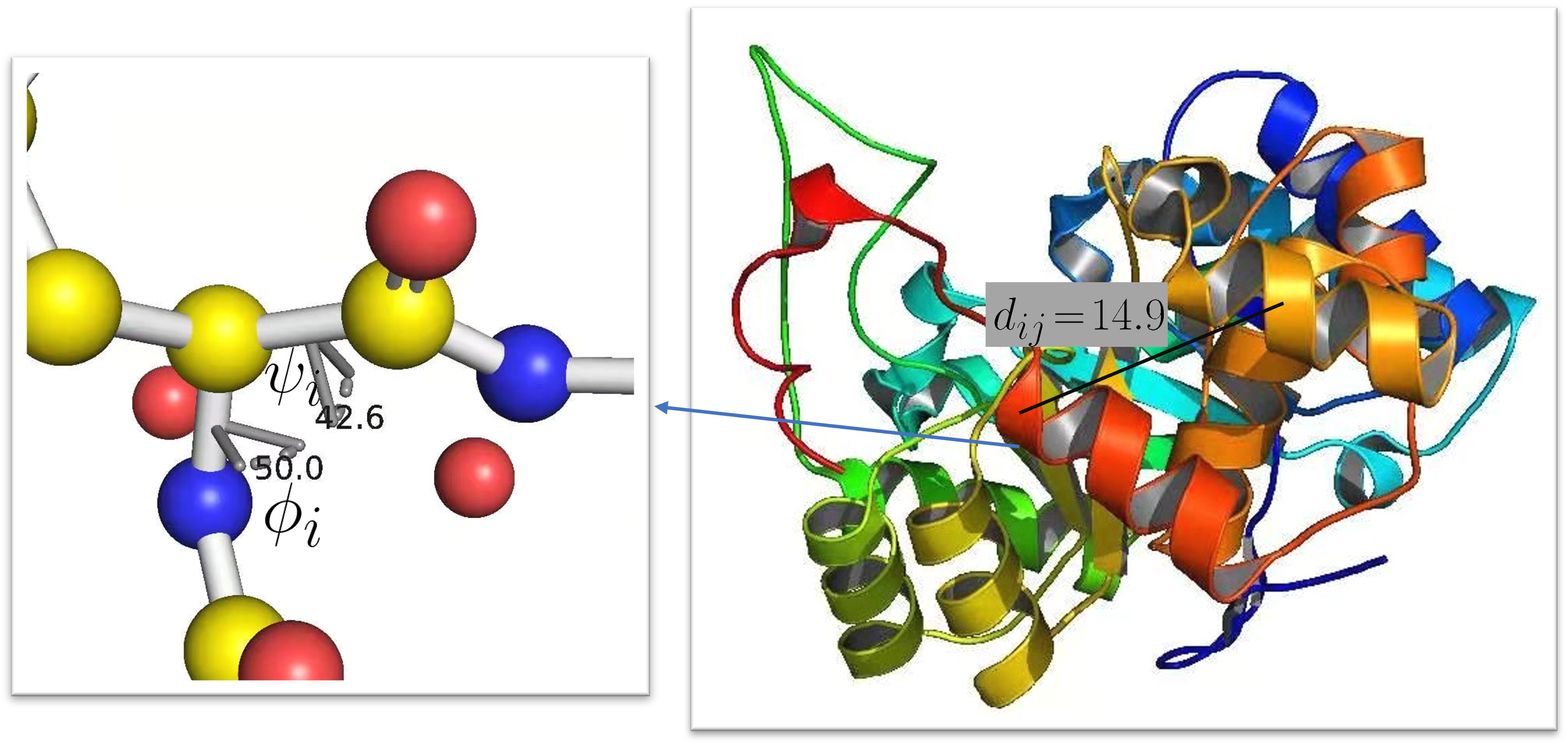}
    \caption{Protein structure.}
    \label{fig:protein_struc}
\end{figure}
\begin{figure}
    \centering
    \includegraphics[width=1.0\columnwidth]{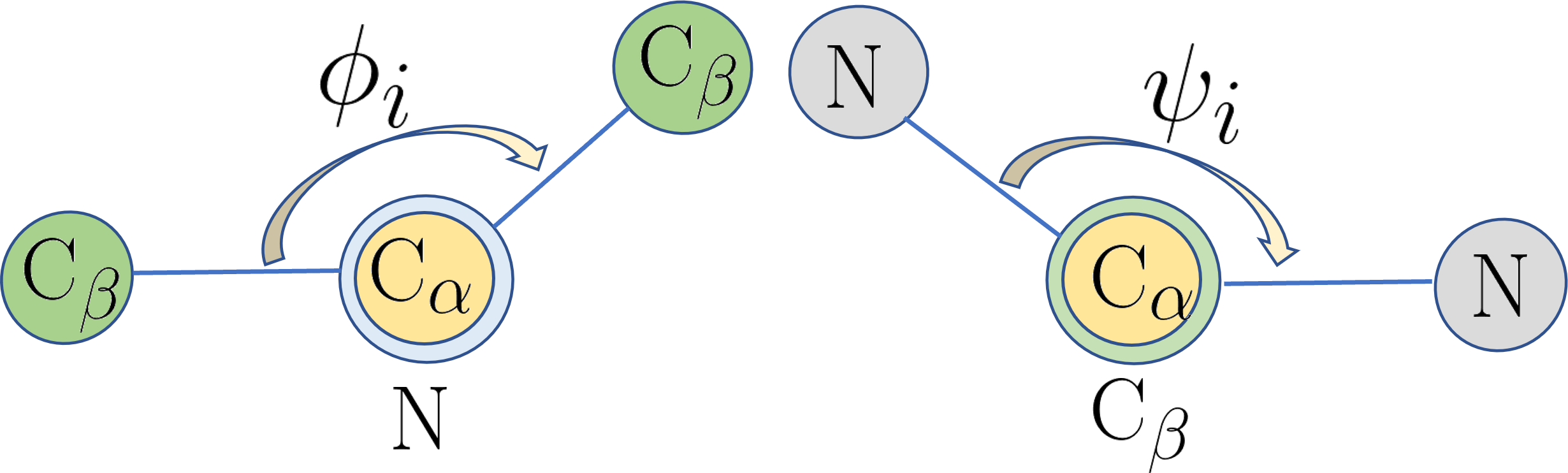}
    \caption{The dihedral angle $\phi_i$ and $\psi_i$.}
    \label{fig:protein_graph}
\end{figure}

\noindent \textbf{Protein structure.}
%\zhoucom{the following part needs to be rewrite
% 1. change different symbols from "Geometric Graph Representation Learning on Protein Structure Prediction"
% 2. clear state the dimension of each symbols
% }
%
%Though a protein may be abstracted as a sequence of amino acids, it represents
%a physical entity with a well-defined three-dimensional structure
%
%\todo{the difference and relation between secondary structure and dihendral angle}
%
%\old{Every protein sequence corresponds to a certain protein structure.}
%\todo{why delete this?}
%
We illustrate an example of protein structure in Figure~\ref{fig:protein_struc}.
As shown in the right {part} of Figure~\ref{fig:protein_struc}, pairwise residue distances provide important structural information {of the protein.} 
%\old{for protein modeling.}
%
%We model protein structure along its backbone.
%
%The protein backbone contains the global structural information and the local structural information.
%
%and .
%
%$d_{ij}$ between $i$ and $j$.
We compute the pairwise residue distance $d_{ij}$ between residue $i$ and residue $j$ as the distance between the corresponding $\rm{C}_{\alpha}$ atoms {on the protein backbone}. 
%We use $d_{ij}$ to represent the alpha carbon \todo{what is alpha carbon} distance between residue $i$ and residue $j$.
%Protein dist
%Global structural information refers to the pairwise distance information
%between residues and 
%
%Let $d_{ij}$ denotes the distance between .
%
Because of the free rotation of the chemical bonds around alpha carbon, distance information alone cannot fully determine protein backbone structure, which necessitates the dihedral angle information.
As shown in the left {part} of Figure~\ref{fig:protein_struc}, the protein backbone consists of consecutive units of $\rm{C}_{\alpha}$–$\rm{CO}$–$\rm{NH}$, and the rotation information around $\rm{C}_{\alpha}$ provides further structural information of {the} protein backbone.
%
%Denote the dihedral angle of the $i_{th}$ residue as $\phi_i$ and $\psi_i$.
%
{A simplified illustration is shown in} Figure~\ref{fig:protein_graph} {where} the two dihedral angles $\phi_i$ and $\psi_i$ capture the rotation of the $\rm{N}$–$\rm{C}_{\alpha}$ bond and the $\rm{C}_{\alpha}$–$\rm{C}_{\beta}$ bond in the residue $i$ respectively. 
% Dihedral angles $\phi_i$ and $\psi_i$ capture 
% rotation of $\rm{N}$–$\rm{C}_{\alpha}$ bond 
%rotation of $\rm{C}_{\alpha}$–$\rm{C}_{\beta}$ bond
%
{In the protein structure, the dihedral angles  $\phi_i$ and $\psi_i$ {are}  two important attributes for each residue $i$.}
%
%local structural information refers to the dihedral angle information. 
%protein structure include secondary structure and tertiary structure.
%
%The protein secondary structure refers to substructures such as $\alpha$-helices and $\beta$-sheets, which are formed by hydrogen bonds between distant amino acids.
%
%
%
%The $\phi$ torsion
%angle describes the rotation of the N – C$\alpha$ bond and the $\psi$ torsion angle describes the rotation of the C$\alpha$ – CO bond.
%
%
%physiochemical properties
%As a result of protein folding, the protein tertiary structure describes the overall shape and function of the protein.
%

%
%In this paper, we use the distance between different amino acids to describe the tertiary structure.

\noindent \textbf{Protein structure as a graph.} 
%
%We model protein from two views: the sequence view and the graph view.
%
%For the sequence view, each protein $\mathcal{S}(\mathcal{V}, \mathcal{E})$ is a sequence of residue linked by the peptide.
%
%Here $\mathcal{V}$ represent the set of $L$ residues in the sequence and $\mathcal{E} \subseteq \mathcal{V} \times \mathcal{V} $ describe the $L-1$ peptide bonds.
%
We model each protein as a graph $\rm{G}(\rm{V}, \rm{E})$ {where $\rm{V}$ denotes the set of nodes in the protein graph and each node represents a residue. Each node $v\in \rm{V}$ has a node feature $\boldsymbol{X_v}$ including the initial residue embedding and the dihedral angle information.  There is an edge $e$ between two nodes in the graph $\rm{G}$ if the pairwise residue distance is smaller than a threshold. \rm{E} represents the set of  edges in the protein graph, and $\boldsymbol{F_e}$ represents the pairwise residue distance information for $e \in \rm{E}$. 
%Here we set the threshold as 7 since ...
} 

% \old{
% with the node feature $X_v$ for $v \in \rm{V}$ and the edge feature $F_e$ for $e \in \rm{E}$ . 
% %\zhou{Similar with protein sequence, $\mathcal{V}$ denotes the set of nodes in the graph which presents the residues. $F$ represents the attribute information of every residue. } 
% %\zhoucom{F is a matrix?}
% %
% %\old{Similarly, $\mathcal{V}$ denotes the amino acids in the graph and $F$ represents the attribute information of every residue.}
% %
% In this paper, $X_v$ includes the initial residue embedding and the dihedral angle information and $F_e$ represents the pairwise residue distance information. 
% %
% %mainly describes the dihedral angles of a certain residue. \zhoucom{use angles as attributes?}
% %
% %\todo{attribute dim}
% %
% %${E}$ represents the edge in the protein graph and $E$ describes the edge attribute such as distance between two alpha carbon atoms. 
% %
% We form the protein graph edges ${E}$ by adding a threshold to the pairwise residue distances.}

\section{The STEPS Framework}
\begin{figure*}
    \centering
    \includegraphics[width=1.00\textwidth]{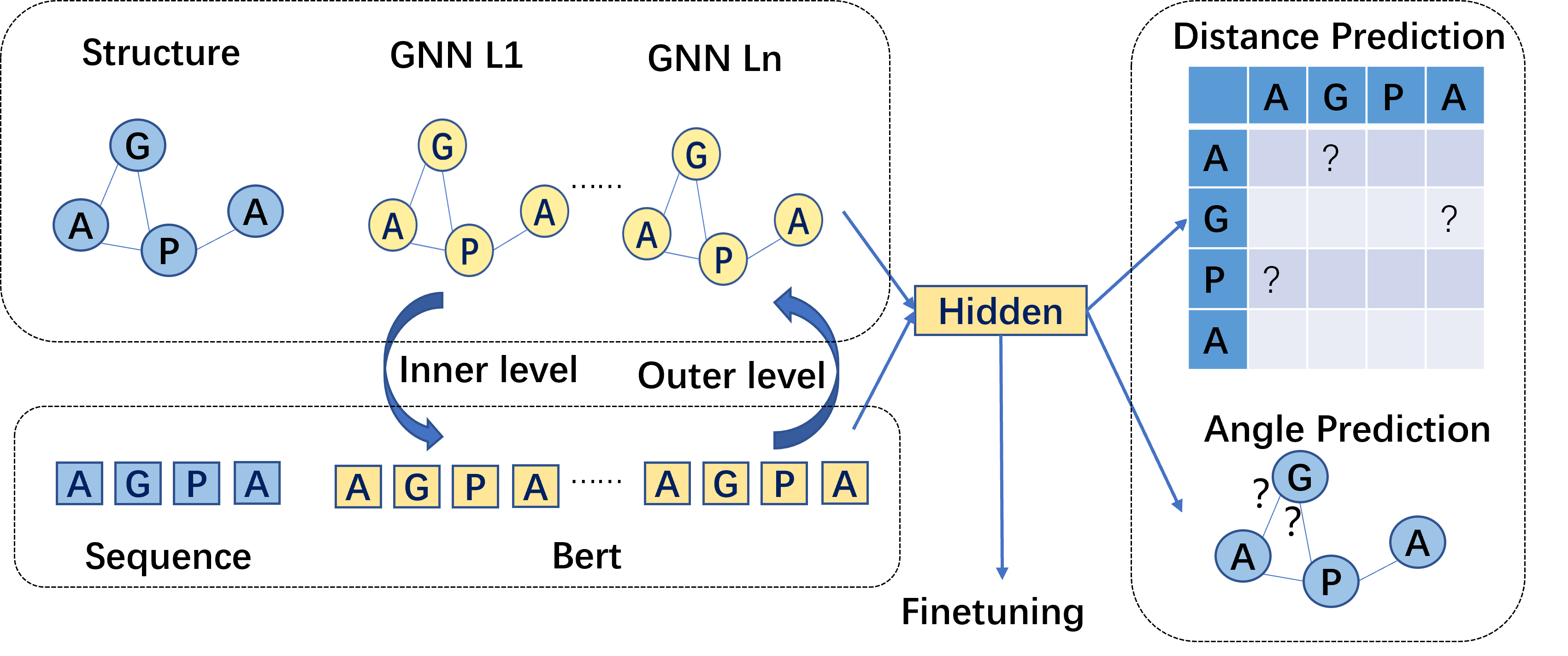}
    \caption{Framework. 
    The GNN model captures protein structural information with two self-supervised tasks: the pairwise distance prediction task and the dihedral angle prediction task. Furthermore, a pseudo bi-level optimization scheme identifies the relation between the protein LM and the GNN model by maximizing the mutual information, which enhances the self-supervised learning.
    }
    \label{fig:flowChart}
\end{figure*}

%before presenting the Structure-aware Protein Language Model~(SPLM) framework.
%
%We first introduce the protein language model which encodes 1D level protein information.
%We aim to pretrain SPLM to yield meaning representation for a certain protein.
%
In this section, we first introduce {a protein modeling method using GNN.} 
%\old{a well-designed graph neural network which can model the  protein structural information explicitly. }
%\old{to incorporate protein structural information into the graph neural network explicitly.}
Second, we present how to use two novel self-supervised tasks to pretrain the GNN model. 
Finally, we introduce the pseudo bi-level optimization scheme.
The overall framework is shown in Figure~\ref{fig:flowChart}.
%
% \old{Two novel self-supervised tasks are proposed to pretrain the graph neural network to preserve the pairwise residue distance information and the dihedral angle information.
% %
% Furthermore we introduce a novel bi-level optimization scheme to \old{build} \zhou{identify} the relation between the LM and the GNN model, and update the graph neural network model under this relation.}

\subsection{Protein Modeling}

%\noindent \textbf{Protein structure.}

%
%Need to write transformer architecture since later analysis.
%
%\begin{equation}\label{equ:score}
%h= \rm{Transformer}(q(S), k(S), v(S)),
%\end{equation}

%\subsubsection{Structure-aware graph neural network}
%\todo{changes to what}
%
%\zhoucom{cannot see where is structure-aware?}
We model a protein structure as a graph and {adopt} a GNN model \cite{xu2018powerful} to {encode} the pairwise residue distance information and the dihedral angle information.
The designed GNN model parameterized by $\boldsymbol{\omega}$ takes as input the protein structural information including {node} features $\boldsymbol{X}$ and edge features $\boldsymbol{F}$, and outputs the node representations and the graph representation.
%\zhoucom{seems this just repeat the GNN?}
%

Denote the node representation for the $i_{th}$ node in the $k_{th}$ layer of the GNN as $\boldsymbol{\rm{h}_i^{(k)}}$.
%
%\old{For the $i_{th}$ node in the $k_{th}$ layer,} 
%\todo{add comma after eq?}
The hidden representation $\boldsymbol{\rm{h}_i^{(k)}}$ is then given by
\begin{equation}
    \boldsymbol{\rm{a}_i^{(k)}} = \rm{AGGREGATE}^{(k)}( e_{iv}\boldsymbol{\rm{h}_{v}^{(k-1)}}\| v \in \mathcal{N}(i))\,,
\end{equation}

\begin{equation}
    \boldsymbol{\rm{h}_i^{(k)}} = \rm{COMBINE}^{(k)}(\boldsymbol{\rm{h}_i^{(k-1)}}, \boldsymbol{\rm{a}_i^{(k)}})\,,
\end{equation}
where $\mathcal{N}(i)$ denotes the neighbors of node $i$ and $e_{iv}$ denotes the feature of the edge between $i$ and $v$. %\zhoucom{the edge between i and v?}
%\old{where we set}
$\rm{AGGREGATE}^{(k)}$ {is} the sum function and %$\rm{AGGREGATE}^{(k)}$
$\rm{COMBINE}^{(k)}$ is a linear layer for feature transformation following~\cite{xu2018powerful} \textcolor{black}{where we feed the sum of the $\boldsymbol{\rm{h}_i^{(k-1)}}$ and $\boldsymbol{\rm{a}_i^{(k)}}$as the input for the linear layer.} %\zhoucom{two same functions?}
We {use} the mean READOUT function to output the graph representation of the protein as:
\begin{equation}
    \boldsymbol{\rm{h}_{G}} = \rm{MEAN}^{(K)}(\boldsymbol{\rm{h}_i^{(K)}} \| i\in {V})\,.
\end{equation}
%
% \old{We adopt a two-layer GNN in this paper.} \zhou{a hyperparamter, seems not necessary here}
%
% \begin{equation}\label{equ:score}
% h_{i} = \rm{GNN}(e_i, \phi_i, \psi_i, (e_j, d_{ij}) | j \in N_i) 
% \end{equation}
%
{Note that} $\boldsymbol{\rm{h}^0}$ refers to the initial node features $\boldsymbol{\rm{X}}$ which mainly include the dihedral angle information and the pretrained node embeddings which serve as initialization.
\textcolor{black}{Specifically, we concatenate the node representation from the pre-trained langauge model and the angle vector into a single input and feed this input into the GNN.}
%and the node embedding from pre-trained language model which serves as initilization.
%
The edge feature $e_{iv}$ refers to the inverse of the square of pairwise residue distance.
%between two connected nodes to represent the edge features.
%
%More specifically, we use the length of the edge to represent the edge feature and the neighbors of some node is averged according the inverse of the square of the length.
%
%\todo{is this convincing?}
%
%We adopt a two layer GNN:
%\begin{equation}
%    hghjj
%\end{equation}
%
%Write down every layer's formula.
%write details in the code

%\subsubsection{Fusion between sequential information and structural information}

%
%The GNN model encodes protein structural information but can not explicitly leverage the sequential information in protein sequence.
%
%To achieve that, we leverage the sequential information from the available pretrained language model. 
%
%Protein language models contain rich 1D protein information.
%
%The model takes the amino acid sequence as input and outputs contextual embedding for every amino acid as the extracted feature representation.
%
%Take \textit{S} as an example.
%
%\textit{S} is passed into the consecutive transformers.
%
% Denote the embedding of the $i_{th}$ residue from the pretrained language model as $h_i^{(s)}$.
% %
% We add the output of pretrained language model into the GNN output,
% %
% %The output of GNN which encodes the 3D level information of protein, is then added to the 1D level representation.
% %
% which can combine the sequential information with the structural information.
To further incorporate the sequential information {of a protein,} we extract protein sequence representation $\boldsymbol{\rm{h}_i^{s}}$ from the protein LM parameterized by $\boldsymbol{\theta}$, and fuse sequential and structural representation for the  residue $i$ as: 
%We fuse the sequential information and the structural information to derive the hidden representation for the residue $i$:
\begin{equation}
    \boldsymbol{\rm{h}_{i}} = \boldsymbol{\rm{h}_i^{s}} + \boldsymbol{\rm{h}_i^{(K)}}\,,
    \label{eq: fusion}
\end{equation}
where $\boldsymbol{\rm{h}_i^{(K)}}$ refers to the final layer hidden representation from the designed GNN model.

\subsection{Self-supervised learning tasks}
%
%\todo{need two pic}
We propose two self-supervised learning tasks to explicitly incorporate the distance information and the angle information into protein modeling. 
The distance prediction task preserves the pairwise residue distance information and the angle prediction task preserves the dihedral angle information.
In this way, the GNN model yields protein representation which well captures the overall protein structural information.
%

%
%We introduce two self-supervised tasks to enhance the learning of node-level and graph-level representations.
%
\subsubsection{Distance prediction task}
%\noindent \textbf{Distance prediction}
%
% \begin{equation}\label{equ:score}
% L_{dis}= \frac{1}{\|V\|^2} \sum_{i, j} -\rm{bin}^T(d_{ij})\times \log (f_{distance}(h_i, h_j))
% \end{equation}
% %
% \textbf{COPY FROM CHEMRL}
% We construct the atomic distance matrix
% for each molecule based on the 3D coordinates of the atoms. Then, we predict the elements in the
% distance matrix, shown in Figure 4(c). We use duv to denote the distance between two atoms u and
% v in the molecule. Note that, for two molecules with the same topological structures, the spatial
% distances between the corresponding atoms could vary greatly. Thus, for a molecule, rather than take
% predicting atomic distance matrix as a regression problem, we take it as a multi-class classification
% problem by discretizing the atomic distances. The loss function is defined as
% %
% where fdistance() is the network predicting the distribution of atomic distances, the bin() function is used to discretize the atomic distance duv into a one-hot vector, and log() is the logarithmic function.
% The task predicting the bond lengths can be seen as a special case of the task predicting the atomic
% distances. The former focuses more on the accurate local spatial structures, while the latter focuses
% more on the distribution of the global spatial structures.
%
The pairwise residue distance determines the overall shape of a protein backbone
and thus determines the function of a protein to a large extent.
%
%The distance between
%It is important to encode the pairwise residue distance information into the GNN model. \zhoucom{revise this sentence later}
%
To this end, we introduce a distance prediction task to encode the pairwise residue distance information into the GNN model.
%

% \old{More specifically, we develop a distance prediction network $\rm{NN}_{dis}()$ which takes the minus of the residue hidden representations and aims to predict the pairwise residue distance.}
%\todo{residues or residues}
{More specifically, we develop a distance prediction network $\rm{NN}_{\boldsymbol{\alpha}_{\rm{dis}}}(\cdot)$ which takes the vector difference between the node hidden representations of residue $i$ and residue $j$ as input, and aims to predict the pairwise residue distance between $i$ and $j$. The intuition for this operation is that the interactions of residues play an important role in determining the diverse functions of protein \cite{cohen2009four}. Therefore,  the residues nearby in the protein backbone should have similar representations.
%
%for protein classification.
}
%
%\todo{to modify}
% \old{\zhou{we hope: 1)  the node representation is more similar if the node are nearby in the physical space; 2) 3D structure of the protein determines the protein function\cite{}. 3) Thus we aim to embed the structure information in the node representation.}}
% \old{The minus operator impose a prior that residues near each other in the physical space are near in the representation space.}
%
{Besides, the numerical scale is quite different in the distance matrix even for the same protein. Therefore, it is more effective to formulate this distance prediction task as a multi-class classification problem instead of a regression problem.}
%
% \old{Besides, we observe the numerical scale is quite different in the distance matrix even for the same protein.
% %
% Therefore, rather than formulating the distance prediction as a regression problem, we formulate this task as a multi-class classification task instead of a regression task. }
%\zhoucom{it is too same with the GEM paper?}
%
We divide the distance into $T$ uniform bins and every bin corresponds to a certain class.
%
%In the experiments, we set $T=30$.
%
In this way, $\rm{NN}_{\boldsymbol{\alpha}_{\rm{dis}}}(\cdot)$ can be written as:
\begin{equation}\label{equ:dij}
d^{'}_{ij} = \rm{NN}_{\boldsymbol{\alpha}_{\rm{dis}}}(\boldsymbol{\rm{h}_i} - \boldsymbol{\rm{h}_j})\,,
\end{equation}
where $d^{'}_{ij} \in \rm{R}^T$ represents the predicted pairwise residue distance distribution  between the residue $i$ and the residue $j$ over $\rm{T}$ classes.
\textcolor{black}{We parameterize $\rm{NN}_{\boldsymbol{\alpha}_{\rm{dis}}}(\cdot)$ as two fully-connected layers with a ReLU activation in the middle.}
%For $\rm{NN}_{\boldsymbol{\alpha}_{\rm{dis}}}(\cdot)$, we set $T$=30 and apply \rm{softmax} to the output logits.
%predicting the distribution of atomic distances
%
%Thus the distance information represents the softmax probability in different bins.
%divide the distance into $T$ bins and every bin
%
%More specifically,
%
%
%Here $d^{'}_{ij} \in D$ is the probability vector of different bins.
%
For a protein, we optimize the Cross Entropy loss among all residue pairs:
%$D^{'}$ and $D$.
%
\begin{equation}\label{equ:ldis}
l_{dis} = \frac{1}{\|V\|^2}\sum_{i, j} -\rm{label}(d_{ij})\log(d^{'}_{ij})\,, 
\end{equation}
where $\rm{label}(d_{ij})$ returns the ground truth one-hot label corresponding to the distance $d_{ij}$.
\subsubsection{Angle prediction task}
{We further propose an angle prediction task for incorporating the dihedral angle information into the GNN model.  The angle prediction task aims to predict the dihedral angles of every residue.  Due to the free rotation of the chemical bonds around the alpha carbon, dihedral angles of residues are of considerable importance since pairwise residue distances alone can not determine the protein backbone structure. Note that the dihedral angles are the attributes of each residue of a protein (instead of between two or more residues).}

% \old{Pairwise residue distance information alone can not determine the protein backbone structure, due to the free rotation of the chemical bonds around alpha carbon.
% %
% The dihedral angles could provide further structural information to the GNN model.
% %
% %to the protein backbone and it is also important to encode the diheral angle information into the GNN model for better protein modeling.
% %\noindent \textbf{Angle prediction}
% %The dihedral angles of every residue are very important geometric parameters and captures the local structure of the protein chain.
% %
% To achieve that, we propose an angle prediction task to predict the diheral angles of every residue.}
%

In particular, we propose an angle prediction network $\rm{NN}_{\boldsymbol{\alpha}_{\rm{ang}}}(\cdot)$ which takes the angle-masked residue representation as input and aims to reconstruct the masked angles.
For a certain protein, 
% \old{we randomly choose $15\%$ of residues.
% %
% %To be specific, 
% We mask the feature of chosen residues}
{we randomly mask the feature of $15\%$ of residues,} and feed the masked protein to the GNN model, which derives the hidden representation of masked residues. 
Note that the dihedral angles are continuous features and we first normalize the angles into [-1, 1].
After that, we adopt the Radial Basis Function to extend the scalar angle information into an angle feature vector, which serves as input to the GNN model.
More specifically, we have
\begin{equation}
    E_{k}({x}) = \rm{exp}(-\gamma\|x-\textcolor{black}{u_j}\|^2)\,,
\end{equation}
where $\gamma$ determines the kernel shape and $\textcolor{black}{\{u_j\}}$ represents the center ranging from -1 to 1.
%
%More specifically, we mask the angles for a portion of residues and denote the output hidden representation of the angle-masked $i_{th}$ as $h_i^m$.
%More specifically, the above process can be written as:
%
% \begin{equation}\label{equ:score}
% h_{i}^{m} = \rm{GNN}(e_i, \rm{mask}, \rm{mask}, (e_j, d_{ij}) | j \in N_i) 
% \end{equation}
%
Denote the final masked representation of the residue $i$ as $\boldsymbol{\rm{h}_i^{m}}$ and then the dihedral angles of the residue $i$ can be predicted as: %\zhoucom{$i_{th}$ residue or  residue $i$?}
\begin{equation}\label{equ:anglepred}
\bar{\phi}_i, \bar{\psi}_i = \rm{NN}_{\boldsymbol{\alpha}_{\rm{ang}}}(\boldsymbol{\rm{h}_{i}^{m}}) \\,
\end{equation}
\textcolor{black}{where we parameterize $\rm{NN}_{\boldsymbol{\alpha}_{\rm{ang}}}(\cdot)$ as two fully-connected layers with a ReLU activation in the middle.}
%
%where $h_i^{(s)}$ is the sequential representation from the protein LM.
%
%We adopt the Mean Squared Error loss for the angle prediction and this can be written as:
%
The Mean Squared Error loss is adopted:
\begin{equation}\label{equ:angleloss}
l_{angle} = \sum_{i \in \mathcal{M}} (\phi_i - \bar{\phi}_i)^2 + (\psi_i - \bar{\psi}_i)^2\,,
\end{equation}
where {$\mathcal{M}$} denotes the set of masked residues.

To sum up, the loss function for the two self-supervised learning can be compactly written as:
\begin{equation}
    \mathcal{L}(\boldsymbol{\theta}, \boldsymbol{\omega}, \boldsymbol{\alpha}) = l_{dis} + l_{angle}\,,
    \label{eq: ssl}
\end{equation}
where $\boldsymbol{\theta}$, $\boldsymbol{\omega}$ and $\boldsymbol{\alpha}$ denote the parameters of the protein LM, the GNN model and the prediction networks, respectively.

\subsection{Pseudo Bi-level Optimization}
Yet, directly fusing representations in Eq.~(\ref{eq: fusion}) can not capture the relation between the sequential information in the protein LM and the structural information in the GNN model.
%
%two levels of information where the 1D level information of the amino acid sequence is closely related to the 3D level information of the protein structure.
%
%To enhance the interplay between the sequential information and the structural information, we propose to build the relation between the pretrained LM and the GNN model.
%
We propose to identify the relation between the protein LM and the GNN model by
maximizing the mutual information between the sequential representation and the structural representation.
%\sout{minimizing the self-supervised loss which reconstructs the protein structure.}
%
We adopt the Jensen-Shannon MI estimator in \cite{nowozin2016f} to estimate the mutual information.
Denote $\boldsymbol{x}$ as a protein sample from $\mathbb{P}$ and $\boldsymbol{\tilde{x}}$ as another protein sample from $\mathbb{\tilde{P}} = \mathbb{{P}} $ , and then we have:
\begin{equation}
    \begin{aligned}
    I(\boldsymbol{\theta}, \boldsymbol{\omega}) =  \mathbb{E}_{\mathbb{P}}[-\rm{sp}(-\rm{T}_{\boldsymbol{\beta}}(h^s_{\boldsymbol{\theta}}(\boldsymbol{x}), h^{(K)}_{\boldsymbol{\omega}}(\boldsymbol{x}))] - \\ \mathbb{E}_{\mathbb{P}\times\mathbb{\tilde{ P}}}[\rm{sp}(\rm{T}_{\boldsymbol{\beta}}(h^s_{\boldsymbol{\theta}}(\boldsymbol{x}), h^{(K)}_{\boldsymbol{\omega}}(\boldsymbol{\tilde{x}}))]\,,
    \label{eq: mututual}
    \end{aligned}
\end{equation}
where $\rm{T}_{\boldsymbol{\beta}}$ denotes the discriminator parameterized by $\boldsymbol{\beta}$ and \rm{sp} is the softplus function.
\textcolor{black}{For the details of $\rm{T}_{\beta}$, we feed the positive and negative examples into a 3-layered fully-connected network with jumping connections and relu activations, and then output the dotproduct of the two representations.}
Then the relation between the sequential parameters $\boldsymbol{\theta}$ and the structural parameters $\boldsymbol{\omega}$ can be identified by maximizing mutual information: 
%The mutual information can be estimated by a neural network as .
%
%
%This can be written as:
%More specifically, we build the relation between $\theta$ and $\phi$ in the inner loop as:
%The connetio can be written as:
\begin{align}
    \boldsymbol{\theta}^{*}(\boldsymbol{\omega}) = \arg \max_{\boldsymbol{\theta}} I(\boldsymbol{\theta}, \boldsymbol{\omega})\,.
    \label{eq: inner}
\end{align}
%
%SSL tasks reconstruct the protein structure.
%which determine
%We build the relation between seq lm and struc GNN by optimizing Eq(11) which reconstruct protein structure.
%
This relation captures the correspondance between the sequential representation and the structural representation for a certain protein~\cite{anfinsen1973principles}.
%
%
%\todo{stress method instead of protein classification}
%\todo{protein protein interaction}
% \sout{This relation is defined by the constraint between protein sequence and protein structure where the protein sequence determines the protein structure~\cite{anfinsen1973principles}.
% %
% %The relation between protein LM and struc GNN is that they both can reconstruct the protein structure by minimizing the SSL loss. 
% %
% Minimizing Eq.~(\ref{eq: inner}) means to adjust $\theta$ to reconstruct the protein structure for a given GNN model parameterized by $\omega$.
% %
% This implies the relation between $\theta$ and $\omega$ is determined by the common protein structure.}
%
%reconstruct protein structure ==  seq and structure consistency
%The SSL task takes the seq info and struc info as input and reconstruct the protein structure, which means it keeps the consistency between seq info and struc info to some extent.
%\zhou{the structure of protein is implicit determined by the sequential amino acids. Therefore, to minimize the Eqn\ref{} means to optimize the $\theta$ to reconstruct the structure of protein while varying structure-aware GNN. }
%
Note that we do not actually update $\boldsymbol{\theta}$ to $\boldsymbol{\theta}(\boldsymbol{\omega})$ in the end, but only
leverage the relation to update the GNN model, which means the final adopted $\boldsymbol{\theta}$ remains the same as that of the protein LM.
%
%model as:
%build the relation between the sequential information and the structural information.
%
%Under this relation, the GNN model is updated as:
%
In this way, the GNN is updated as:
\begin{equation}
\boldsymbol{\omega}^{'} = \arg \min_{\boldsymbol{\omega}} \mathcal{L}(\boldsymbol{\theta}(\boldsymbol{\omega}), \boldsymbol{\omega}, \boldsymbol{\alpha})\,,
\label{eq: outer}
\end{equation}
which could better exploit the sequential information in the protein LM.

This can be formulated as a bi-level optimization problem:
\begin{alignat}{2}
\min_{\boldsymbol{\omega}, \boldsymbol{\alpha}} \quad & \mathcal{L}(\boldsymbol{\theta}(\boldsymbol{\omega}), \boldsymbol{\omega}, \boldsymbol{\alpha}) \,,& \\
%\label{eq: inner1}
\mbox{s.t.}\quad & \boldsymbol{\theta}^*(\boldsymbol{\omega}) = \mathop{\arg\max}_{\boldsymbol{\omega}} I(\boldsymbol{\theta}, \boldsymbol{\omega}) &\,.
%\label{eq: outer1}
\end{alignat}
%
%where Eq.(\ref{eq: inner1}) defines the outer level task and Eq.(14) defines the inner level task.
%
Different from the traditional bi-level optimization, we do not update $\boldsymbol{\theta}$ in the end similar to~\cite{wang2018dataset} so we name this scheme as pseudo bi-level optimization.
%
%In the inner level, we identify the relation between $\theta$ and $\omega$.
%
The inner level can be solved approximately by a gradient ascent step:
\begin{equation}
\boldsymbol{\theta}(\boldsymbol{\omega}) = \boldsymbol{\theta} + \eta * \frac{\partial I(\boldsymbol{\theta}, \boldsymbol{\omega})}{\partial \boldsymbol{\theta}}.
\end{equation}
%\todo{low level}
%
%In the outer level, the structural information $\omega$ is updated under the inner level relation.
%
%Similarly, this optimization process can be simplified into a gradient descent step as:
Similarly, the outer level can be solved as:
\begin{alignat}{2}
\boldsymbol{\omega}^{'} = \boldsymbol{\omega} - \eta^{'} * \frac{\partial \mathcal{L}(\boldsymbol{\theta}(\boldsymbol{\omega}), \boldsymbol{\omega}, \boldsymbol{\alpha})}{\partial \boldsymbol{\omega}}\,, \\
\boldsymbol{\alpha}^{'} = \boldsymbol{\alpha} - \eta^{'} * \frac{\partial \mathcal{L}(\boldsymbol{\theta}(\boldsymbol{\omega}), \boldsymbol{\omega}, \boldsymbol{\alpha})}{\partial \boldsymbol{\alpha}}\,.
\end{alignat}

%% file: Sec3_experiment.tex
\section{Experiments}
{In this section, we first introduce the pretraining settings including the datasets and the training details. Second, we evaluate STEPS on three supervised downstream tasks and compare STEPS with existing SOTA methods. At last, we conduct ablation studies to verify the effectiveness of different components in our method STEPS.}

\subsection{Pretraining settings}
\noindent \textbf{Datasets.}
% We merge two {datasets} from the Deeploc dataset \cite{almagro2017deeploc} and the Enzyme dataset~\cite{hermosilla2020intrinsic}, and perform pretraining on the merged dataset.
% %
% For the Deeploc dataset, we acquire the available protein structures from the alphafold protein database~\footnote{https://alphafold.ebi.ac.uk/}.
% %
% For the Enzyme dataset, we use PDB files in the Protein Data Bank~\cite{berman2000protein} to {obtain} protein structures.
% %
% Besides, we exclude proteins with a sequence length of more than 400 residues, which results in a total of around 40,000 protein {structures} for pretraining.
%
\textcolor{black}{We sample an independent set from the alphafold protein database~\footnote{https://alphafold.ebi.ac.uk/} and remove protein sequences which have more than $25$\% sequence similarity with the test proteins, forming a size-$40000$ pretraining set.}
%we have removed the test proteins from the pretraining dataset and sample an independent set of the same size from the alphafold2 database to form a pretraining set.

%\noindent \textbf{Model.}
\noindent \textbf{Training details.}
For the GNN model, we set the {dimension of} hidden representation as $1280$ and the layer number as $2$ in our experiments.
{The threshold to determine whether there is an edge between two residues is set as 7 \AA~which is consistent with previous study \cite{xia2021geometric}.}
%\old{We define the edge between two residues by a threshold of 7 Å.}
%
%We use a fully-connected layer to 
%We parameterize both $\rm{NN}_{\boldsymbol{\alpha}_{\rm{dis}}}(\cdot)$ and $\rm{NN}_{\boldsymbol{\alpha}_{\rm{ang}}}(\cdot)$ as two fully-connected layers with a ReLU activation in the middle.
%
For $\rm{NN}_{\boldsymbol{\alpha}_{\rm{dis}}}(\cdot)$, we set $T$=30 and apply \rm{softmax} to the output logits.
Besides, we adopt the available protein BERT model in \cite{elnaggar2020prottrans} as the pretrained protein language model.
%
%For $\rm{T}_{\beta}$ in mutual information estimation, we feed the positive and negative examples into a 3-layered fully-connected network with jumping connections and relu activations, and then output the dotproduct of the two representations.
%For the protein language model, we adopt one auto-regressive models (XLNet) and one auto-encoder models (BERT).
%
%
%\noindent \textbf{Training details.}
%
%We adopt the cosine learning decay policy with 

We use the cosine learning rate decay schedule for a total of 10 epochs for pretraining.
We set the learning rate for the GNN model as $1e^{-3}$ and the learning rate for the protein LM as $5e^{-5}$ in the pseudo bi-level optimization scheme.
The Adam optimizer is adopted to update the GNN parameters with $\beta_1 = 0.9$ and $\beta_2 = 0.999$.
%bilevel pretraining.

\subsection{Finetuning}

\noindent \textbf{Downstream tasks.}
We finetune the pretrained model on three downstream tasks: the binary classification into membrane/non-membrane proteins, the location classification into $10$ cellular compartments~\cite{almagro2017deeploc}, and the
enzyme-catalyzed reaction classification~\cite{hermosilla2020intrinsic} into $384$ Enzyme Commission (EC) numbers.
{Hereafter} we denote the three tasks as C2, C10, and C384 respectively for convenience.
\textcolor{black}{The train/test sizes of C2, C10, and C384 are $2221/568$, $3874/988$, and $15001/2799$ respectively. For the binary classification, the membrane/non-membrane protein of the train is $1123/1098$ and the membrane/non-membrane protein of the test is $297/271$.}
The performance is evaluated as the mean accuracy~(acc) following the setting in ~\cite{hermosilla2020intrinsic}.
%We mainly use two datasets Deeploc and Enzyme, and use three tasks: Deeploc Q2, Deeploc Q10 and Enzyme Classification as our evaluation protocol. 

\noindent \textbf{Baselines.}
We compare STEPS with two {groups} of baselines: methods without {and with pretraining.}
Methods without pretraining include:
\begin{itemize}
    \item Blast~\cite{radivojac2013large}: a sequence in the test set receives labels from all labeled sequences in the training set and the prediction is obtained as the highest one. Similar to \cite{gligorijevic2021structure}, we remove all training sequences with an E-value threshold $1e$-$3$ to prevent label transfer from homologous sequences.
    \item IEConv~\cite{hermosilla2020intrinsic}: {it} introduces a novel convolution operator and hierarchical pooling operators to model different particularities for a protein.
    %\old{\item DeepFRI~\cite{gligorijevic2021structure}: this method adopts the Graph Convolutional Network to predict protein functions by sequence features and structure features.}
\end{itemize}
%
%\old{We consider two methods with pretraining:}
{Methods with pretraining are:}
%For methods with pretraining,  we pretrain the designed GNN model with the same pretraining setting
%consider the pretrained protein bert model and train  consider  consider only use the language model or only the GNN.
\begin{itemize}
    \item Pre-LM~\cite{elnaggar2020prottrans}: {it} adopts the protein BERT model pretrained on Uniref100 and adds an fully-connected layer with tanh activation as the head for finetuning.
    The head takes the mean pooling over residue representations as input and outputs scores. 
    \item DeepFRI~\cite{gligorijevic2021structure}: this method adopts the Graph Convolutional Network (GCN) to predict protein functions by leveraging structural features. It also adopts a pretrained language model to obtain the residue embedding as the input of the GCN model.
    %  \old{\item Pre-GNN: we pretrain the designed GNN model without considering the pretrained LM and add a layer on GNN for finetuning similar to LM.}
    \item STEPS-w/oLM: we pretrain the GNN model in STEPS without considering protein LMs. We add a layer on GNN for finetuning. \textcolor{black}{The initial node representation for GNN is the concatenation of the one-hot encoding of amino acid and the dihedral angle vectors.}
    %finetune on the head similar to the LM method.
\end{itemize}
%
%For methods with pretraining, we further consider two cases: 
%
For the proposed STEPS, we finetune both the GNN model and the linear head.
{Besides we use STEPS-H to denote the STEPS with only finetuning the linear head.}
%\old{Besides, we denote only finetuning the linear head as STEPS-H.}
%and finetuning the GNN model and the linear head as SPM-B. 

%Traditional methods(evolutionary information);
%Embedding from pretrained LM;
%Embedding from normally finetuning pretrained LM;
%Embedding from bi-level finetuning pretrained LM;

% \begin{itemize}
%     \item Without pretraining: GNN
%     \item With pretraining: LM; GNN
%     %\item Variants of our method: , with only one pretrain task;
% \end{itemize}

\noindent \textbf{Training details.}
For all methods and all datasets, we adopt a cosine learning rate decay with an initial learning rate $1$e-$4$ and train the models for $5$ epochs with the Adam optimizer for a fair comparison. 
%
%See Appendix for details.

%
% loc

% before train 11085 after train 9065

% before test 2773 after test 2313

% water

% before train 6913 after train 5271

% before test 1749 after test 1360

% \begin{figure}
%     \centering
%     \includegraphics[width=0.45\textwidth]{Figures/ablation.png}
%     \caption{Ablation studies}
%     \label{fig:ablation}
% \end{figure}

\subsection{Result Analysis}
\vspace{-2pt}
\begin{table}
\centering
% \scriptsize
\caption{Experimental Results on C2 for Comparison.}
\scalebox{.65}{
\begin{tabular}{ccccccccc}
\toprule
Method & Blast& IEConv &  DeepFRI & Pre-LM & STEPS-w/oLM & STEPS-H & STEPS \\
\midrule
Acc(\%) & $65.14$ &$62.15$ &\underline{$88.38$} & $58.00$ & $77.82$&$87.68$&$\textbf{89.61}$\\
\bottomrule
\end{tabular}
 }
\label{table: Deeploc-Q2}
\end{table}

%\textbf{unsupervised fold classification}

\begin{table}
\centering
% \scriptsize
\caption{Experimental Results on C10 for Comparison.}
\scalebox{.65}{
\begin{tabular}{ccccccccc}
\toprule
Method & Blast& IEConv &  DeepFRI & Pre-LM & STEPS-w/oLM & STEPS-H & STEPS \\
\midrule

Acc(\%) & $31.78$ & $30.99$ &{$\underline{69.23}$} &$35.00$ & ${42.51}$&${69.84}$&$\textbf{71.05}$\\
\bottomrule
\end{tabular}
 }
\label{table: Deeploc-Q10}
\end{table}

\begin{table}
\centering
% \scriptsize

%\todo{rerun}
\caption{Experimental Results on C384 for Comparison.}
\scalebox{.65}{
\begin{tabular}{cccccccc}
\toprule
Method & Blast& IEConv &  DeepFRI & Pre-LM & STEPS-w/oLM & STEPS-H & STEPS \\
\midrule
Acc(\%) & $12.83$ & $\underline{28.70}$ &{$15.72$} &$1.32$ & $3.89$ & ${50.13}$ & $\textbf{65.38}$\\
\bottomrule
\end{tabular}
 }
\label{table: Spe-Cls}
\end{table}

As shown in Table~\ref{table: Deeploc-Q2}, Table~\ref{table: Deeploc-Q10} and Table~\ref{table: Spe-Cls}, we report the best results in bold and mark the second best results~(excluding STEPS-H) among two groups of baselines by underlines.
First, we can observe that STEPS has consistent gains over all comparison methods in {the} three downstream tasks.
More specifically, compared with the second best results, STEPS achieves $1.39$\% relative gain in C2, $2.63$\% relative gain in C10 and $36.68$\% performance gain in C384, which proves the effectiveness of STEPS.
Note that STEPS performs better than STEPS-H, which means further finetuning the GNN model on a specific task yields better representation.
{It is worth noting that STEPS significantly outperforms its baselines in C384. A potential reason is that protein structure primarily determines the specific binding sites of an enzyme. Therefore, as the first method to incorporate the structural information into protein pretraining, STEPS performs much better than other methods on the enzyme-catalyzed reaction classification task~(i.e. C384).}
\textcolor{black}{ We conduct additional experiments where the test set structures are from Alphafold instead of the PDB database on C384, and the results are very similar. The STEPS accuracy on C384 is 65.38\% when using the structures from Alphafold and is 65.74\% when using the PDB database.}
Furthermore, we can observe that Pre-LM and STEPS-w/oLM perform worse than STEPS, which verifies the necessity of structural information and  sequential information for protein pretraining.
At last, we can observe STEPS-w/oLM performs better than Pre-LM by $19.82$\% in C2, $7.51$\% in C10 and $2.57$\% in C384, which indicates structural information is more important than sequential information for {protein representation learning.} 
%\old{protein pretraining.}
%
%Last but not least, we observe that BLAST serves a good baseline across three tasks which demonstrates the sequence identity information alone contains much useful information for protein classification.\zhoucom{is this sentence necessary?}

%although BLAST only leverages the sequence identity information.

% Following \cite{rao2019evaluating}, we adopt one evolutionary understanding task(i.e., remote homology detection) and two protein engineering tasks(i.e. fluorescence landscape prediction and stability landscape prediction) to evaluate the performance of pretrained LM.
% \subsubsection{Remote Homology Detection}
% \subsubsection{Fluorescence Landscape Prediction}
% \subsubsection{Stability Landscape Prediction}

% Design SSL tasks
% Chem
% \begin{itemize}
%     \item node-level: 21
%     \item context
%     \item graph-level
% \end{itemize}
% Struc
% \begin{itemize}
%     \item node angle
%     \item bond
%     \item res dis

%\subsection{Embedding Understanding}

\subsection{Ablation Studies}
%

% \begin{figure}
%     \centering
%     \includegraphics{Figures/ablation.png}
%     \caption{Caption}
%     \label{fig:my_label}
% \end{figure}

\begin{figure}
    \centering
    \includegraphics[width=1.0\columnwidth]{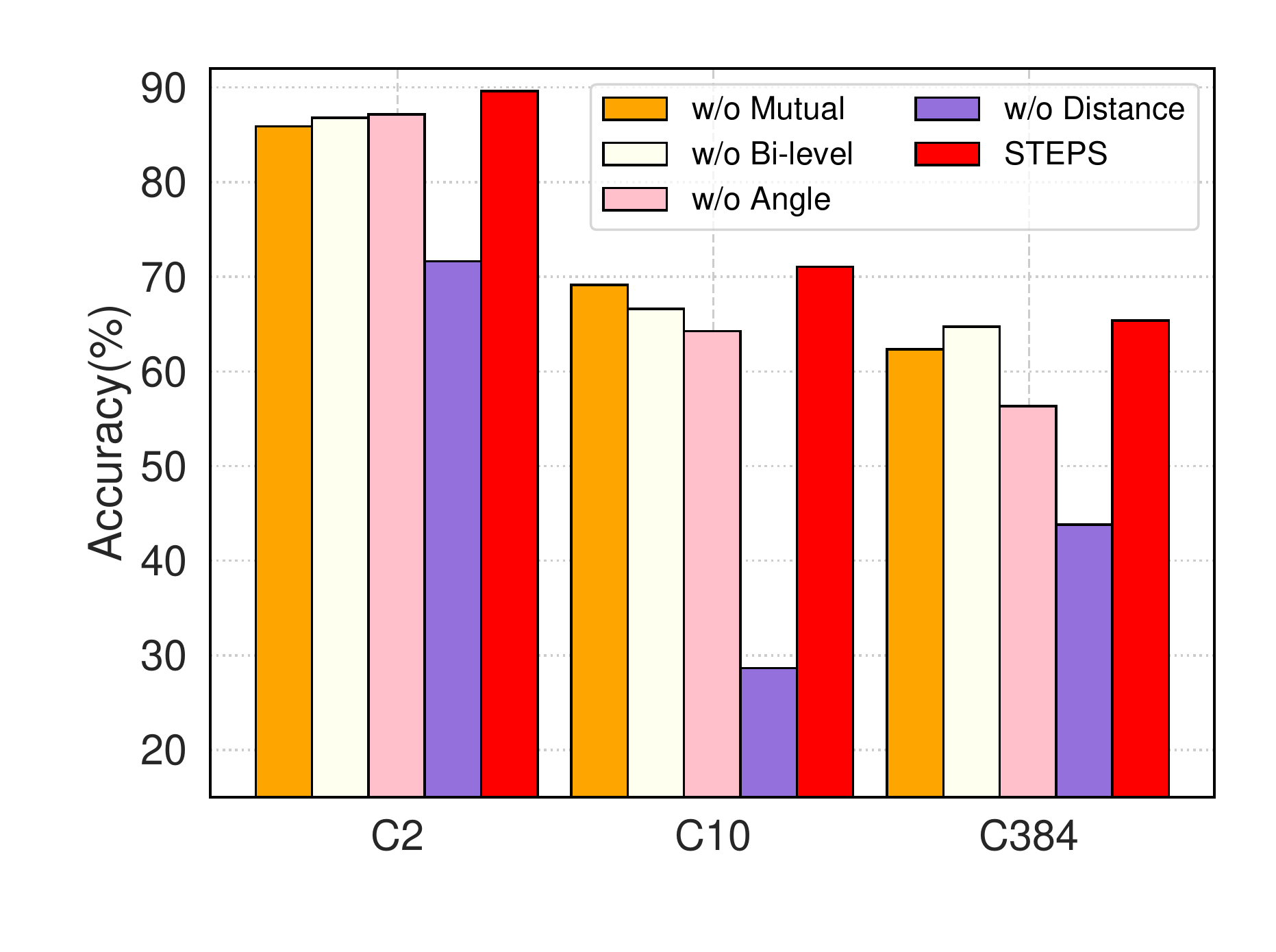}
    \vspace{-30pt}
    \caption{Ablation studies.}
    \label{fig: ablation1}
\end{figure}

In this section, we conduct ablation studies to verify the effectiveness of different components in STEPS.

%
%We aim to answer two research questions:
%\begin{itemize}
%    \item RQ1:
%    \item RQ2: 
%\end{itemize}
%We further conduct ablation study to verify the design component of SPLM.
%
%Some of the following may not be ablation.
\subsubsection{Mutual information}
We first remove the mutual information from STEPS, denoted as w/o Mutual, and only use the self-supervised learning losses.
As shown in Figure~\ref{fig: ablation1}, this removal leads to decreases on all three tasks, which demonstrates the importance of the mutual information. 

% the performances of three tasks decrease a lot, which indicates the importance of the mutual information. 
% Bi-level: 88.91 70.55 66.42

% Joint: 87.15, 68.62, 64.95

% Bi-level wo global(distance): 72.71,  25.81, 38.12

% Bi-level wo local(angle): 87.15, 34.11, 59.02

% Remove mutual: 81.87, 68.62, 63.85

\subsubsection{Pseduo bi-level optimization}
To verify the {effectiveness} of the pseudo bi-level optimization, we remove this part from STEPS and only optimize the GNN model, which is denoted as w/o Bi-level.
%without bi-level pretraining
%; compute the mutual information between 1D level representaion and 3D level representation.
Besides, we consider alternate optimization and joint optimization between the protein LM and the GNN model.
We find alternate optimization and joint optimization {do not} perform {well and are even much worse than without these optimizations, so we do not report these results here.} 
%\old{badly and even reduces to random predictions so we do not report results here.}
%
The reason may be the protein LM collapse during updating~\cite{dodge2020fine}.
As shown in Figure~\ref{fig: ablation1}, STEPS {consistently} outperforms {STEPS} w/o Bi-level {in all tasks}, which verifies the effectiveness of the proposed pseduo bi-level optimization.
%
%Furthermore, alternate optimization performs the worst and the reason may be that changing the pretrained language model parameters leads to collapse.

% \begin{table}
% \centering
% % \scriptsize

% \caption{Bilevel pretraining.}
% \scalebox{.88}{
% \begin{tabular}{cccc}
% \toprule
% Task & Deeploc Q2 & Deeploc Q10 & Enzyme cls \\
% \midrule
% SPM w/o Bil & \underline{$85.56$} & $67.51$ &$62.38$\\
% \midrule
% SPM & \underline{$89.26$} & $70.64$ &$66.85$\\
% \bottomrule
% \end{tabular}
%  }
% \label{table: bi-level}
% \end{table}

%\subsubsection{Is the designed ssl task necessary?}
\subsubsection{Self-supervised learning tasks}
{We then demonstrate the effectiveness of the two self-supervised learning tasks:}
%\old{Furthermore, we want to verify the effectiveness of the two self-supervised tasks:} 
the pairwise residue distance prediction task and the dihedral angle prediction task.
We remove the angle prediction task from STEPS and denote it as w/o Angle.
Similarly, we remove the distance prediction task from STEPS and denote it as w/o Distance.
%
%the proposed self-supervised task from SPLM and only keep one self-supervised task for pretraining. 
%with only one pretrain task, the tradeoff of two tasks;
%
As shown in Figure~\ref{fig: ablation1}, removing either task leads to {noticeable} performance degradation, which proves the necessity of both self-supervised learning tasks.
{Moreover,} we observe that STEPS w/o Distance results in $15.50$\% performance {decline} in C2, $35.63$\% performance {decline} in C10 and $12.54$\% performance {decline} in C384 compared with STEPS w/o Angle.
{This phenomenon indicates that the pairwise residue distance information plays a more important role than the dihedral angle information in protein modeling. }
\textcolor{black}{We follow the work~\cite{fang2022geometry}\cite{xia2021geometric} to model the distance prediction as a multi-class classification problem rather than a regression problem. The regression formulation leads to 3.87\% performance decline in C2, 1.01\% performance decline in C10 and 2.11\% performance decline in C384.}

%\old{This means the pairwise residue distance information plays a more important role than the dihedral angle information in protein structure modeling.}

% \begin{table}
% \centering
% % \scriptsize

% \caption{Ssl task pretraining.}
% \scalebox{.88}{
% \begin{tabular}{cccc}
% \toprule
% Task & Deeploc Q2 & Deeploc Q10 & Enzyme cls \\
% \midrule
% w/o Angle & \underline{$85.74$} & $67.61$ &$62.81$\\
% \midrule
% w/o Distance & \underline{$49.47$} & $17.91$ &$43.37$\\
% \midrule
% SPM & \underline{$89.26$} & $70.64$ &$66.85$\\
% \bottomrule
% \end{tabular}
%  }
% \label{table: ssl}
% \end{table}

%\subsubsection{Is different lm model important?}
%

%different pretrained lm(Bert, TXL, etc); Bert: auto-encoder model;Transformer-XL: auto-regressive language model
%Abalation study

% xlnet lm
% enzyme 0.1854
% loc 0.3421
% water 0.5106

% xlnet gnn
% loc 0.4312
% water 0.7975
% enzyme 0.0493

% \begin{table}
% \centering
% % \scriptsize

% \caption{LM choice.}
% \scalebox{.88}{
% \begin{tabular}{cccc}
% \toprule
% Task & Deeploc Q2 & Deeploc Q10 & Enzyme cls \\
% \midrule
% Bert & \underline{$00.00$} & $00.00$ &$00.00$\\
% \midrule
% TXL & \underline{$00.00$} & $00.00$ &$00.00$\\
% \midrule
% Albert & \underline{$00.00$} & $00.00$ &$00.00$\\
% \bottomrule
% \end{tabular}
%  }
% \label{table: lm}
% \end{table}

% \end{itemize}

%% file: Sec4_relatedwork.tex
\section{Related Work}
\subsection{Protein Representation Learning}
Protein representation learning methods are mainly classified into two categories: sequence-based methods and structure-based methods.
Sequence-based methods model a protein via its one-dimensional amino acid sequence.
For example, \cite{hou2018deepsf} adopt one-dimensional convolutional neural networks to derive hidden representation for classification.
%
%LSTM
%Transformer
Structure-based methods consider the three-dimensional (3D) structure of proteins.
For example, \cite{townshend2019end} leverage 3D convolutional neural networks for protein quality assessment and protein contact prediction.
\cite{hermosilla2020intrinsic} propose novel convolutional operators and pooling operators to model the primary, secondary, and tertiary structure effectively, {which} demonstrates strong performance on protein function prediction tasks. 
\cite{gligorijevic2021structure} leverage the LSTM model to encode the {protein} sequence and the GCN model to encode the {protein} tertiary structure for function prediction. 
\cite{somnath2021multi} connect protein surface to structure modeling and sequence modeling {where} the learned
representation achieves {good} performance on several downstream tasks.
%Some research work also propose to model the protein surface.

\subsection{Protein Pretraining}
%\todo{discuss some work implicitly use protein structure for pretraniing}
%
%\old{\noindent Inspired by the recent process of self-supervised learning in natural language processing, a few recent work perform pretraining on protein sequences.}
There are a few studies to perform pretraining on protein sequences
~\cite{bepler2019learning,zhou2023protein,wang2023on}. 
%Recently, language modeling for protein sequences has received lots of attention.
%
\cite{bepler2019learning} {propose to} train an LSTM on protein sequences, which {could} implicitly incorporate {structural} information from the global structural similarity between proteins and the contact maps for individual proteins, while STEPS uses novel self-supervised tasks {to} explicitly model protein structure.
\textcolor{black}{Our distance prediction task and the contact prediction task in \cite{bepler2019learning,bepler2021learning} can both incorporate the distance information into the learned protein representation. The difference is that the distance prediction task models the distance as a multi-class classification and this can explicitly consider more protein structural information compared with the binary classification of the contact prediction.}
%
%\cite{rao2019evaluating} evalua set of five biologically relevant semi-supervised
%learning tasks spread across different domains of protein biology.
%\cite{rao2019evaluating} propose a panel of benchmarks for protein language model evaluation, which boost the development of this area.
%
\cite{rives2021biological} is the first to model protein sequences with self-attention, and the learned representation of the pretrained language model contains the protein information of structure and function.
%\zhou{The representation from this pretrained language model demonstrates that it can learn useful protein information for determining protein structure and functions.}
%
%
\cite{elnaggar2020prottrans} {try to} train auto-regressive language models and auto-encoder models on large {datasets}, and validate the feasibility of training big language models on proteins.
%
%\cite{rao2021msa} takes as input a set of sequences of multiple sequence alignment and propose the MSA transformer, which achieves state-of-the-art performances across many tasks.
%
\cite{rao2020transformer,vig2020bertology}
study the transformer attention maps from the unsupervised learned language model and uncover the relationship between the attention map and the protein contact map.
\cite{fang2022geometry} design similar self-supervised learning tasks for molecules while STEPS consideres different information including the pairwise residue distance information and the dihedral angle information for protein modeling.
Besides, STEPS models protein from both sequence and structure views while \cite{fang2022geometry} model molecule from only the structure view.
A concurrent work of STEPS~\cite{zhang2022protein, zhang2023enhancing} adopts a well-designed GNN for protein pre-training.
A future direction may be adopting the pre-trained protein models for protein optimization~\cite{chen2023bidirectional}.

% \cite{alley2019unified}
% \cite{heinzinger2019modeling}

\subsection{Bi-level Optimization}
%\subsection{Language Model Fine-tuning}
%\cite{howard2018universal}
%\subsection{Protein Classification}
%The species is the only taxonomic category that exists in nature. 
Bi-level optimization is a special kind of optimization problem where one level of problem is embedded in the other level.
Bi-level optimization has been widely used in the deep learning community due the hierarchy problem structure in many {applications}~\cite{hospedales2020meta,chen2022gradient,chen2021generalized,can2022bidirectional,chen2022unbiased} including neural architecture search, instance weighting, initial condition, learning to optimize, data augmentation, etc.
In this paper, similar to \cite{wang2018dataset}, we develop a pseudo bi-level optimization scheme to identify the relation between the sequential information in the protein LM and the structural information in the GNN model, which can {help} exploit the sequential information in the protein LM. 

%In this paper, to enhance the interaction between the LM and GNN model, we propose a bi-level learning framework where we build the connection between the LM parameters and the GNN model.
%
%In this way, the structural information is not only updated through the GNN model but also via the LM model, which could better exploit the knowledge in the pretrained language model.

%GNN model is  updated under this relation.
%
%we propose a bi-level learning framework where we build the connection between the LM parameters information and the GNN parameters.
%information via optimization methods.
%

%% file: Sec5_conclusion.tex
\section{Conclusion}

{In this paper, to effectively capture protein structural information, we investigate a novel structure-aware self-supervised protein learning approach. 
%Specially, we propose to utilize both the protein structure information and the  pretrained sequential protein language model to improve the protein representation learning for classification. 
Along this line, two novel self-supervised learning tasks on a GNN model is adopted to capture the pairwise residue distance information and the dihedral angle information, respectively. Also, to leverage the pretrained sequential protein language model to further improve the representation learning, we propose a pseudo bi-level optimization scheme to transfer the knowledge of the protein LM to the GNN model. Finally, the experimental results on several benchmarks for protein classification show the effectiveness and the generalizability of our method STEPS.
%In future, we plan to study how to apply the protein representation learning to other protein related problems like protein design and protein-ligan interaction.
%Potential future work may apply the learned protein representation to the area of protein design, protein-ligand interaction, etc.
}